\documentclass{article}
\usepackage{ijcai20}

\usepackage{times}
\usepackage{soul}
\usepackage{url}
\usepackage[small]{caption}
\usepackage{graphicx}
\usepackage{amsmath}
\usepackage{amsfonts}
\usepackage{amsthm}
\usepackage{booktabs}
\usepackage{algorithm}
\usepackage{float}
\usepackage{algorithmic}
\DeclareMathOperator*{\argmax}{argmax} 

\begin{document}

\title{Optimal Use of Multi-spectral Satellite Data with Convolutional Neural Networks}

\author{Sagar Vaze
\and
Conrad James Foley \and Mohamed Seddiq \and Alexey Unagaev \and Natalia Efremova
\\
{\tt\small (sagar,james,mohamed,alexey,natalia)@deepplanet.ai}
}

\maketitle

\begin{abstract}
   
      The analysis of satellite imagery will prove a crucial tool in the pursuit of sustainable development. While Convolutional Neural Networks (CNNs) have made large gains in natural image analysis, their application to multi-spectral satellite images (wherein input images have a large number of channels) remains relatively unexplored. In this paper, we compare different methods of leveraging multi-band information with CNNs, demonstrating the performance of all compared methods on the task of semantic segmentation of agricultural vegetation (vineyards). We show that standard industry practice of using bands selected by a domain expert leads to a significantly worse test accuracy than the other methods compared. Specifically, we compare: using bands specified by an expert; using all available bands; learning attention maps over the input bands; and leveraging Bayesian optimisation to dictate band choice. We show that simply using all available band information already increases test time performance, and show that the Bayesian optimisation, novelly applied to band selection in this work, can be used to further boost accuracy.
   
\end{abstract}


\section{Introduction}

The analysis of satellite imagery will play a critical role in the pursuit of the UN's sustainable development goals (SDGs) \cite{united2015transforming}. Analysis of this data has been proposed for applications from identifying clean water sources; to surveillance of human trafficking; to estimating the economic well-being of cities in developing countries \cite{dig_globe_report}. Recent work has shown that earth observation data can partially or majorly contribute to a quarter (55 of 207) of the SDG indicators \cite{andries2019seeing}. Furthermore, the global Geographic Information System (GIS) market is expected to be worth over 14 billion dollars by 2025, with an estimated 12\% year-on-year growth \cite{research_and_markets2020}. 

Meanwhile, convolutional neural networks (CNNs) have recently made large advances in most natural image analysis tasks. However, differently to `RGB' natural images, a key feature of satellite imagery is its multi-spectral nature, in which information on radiation of multiple wavelengths is recorded and can be leveraged to gain insight. It is thus important that models which operate on satellite imagery are designed to optimally utilise the multi-band information.

\begin{figure}
    \centering
    \includegraphics[width=\linewidth]{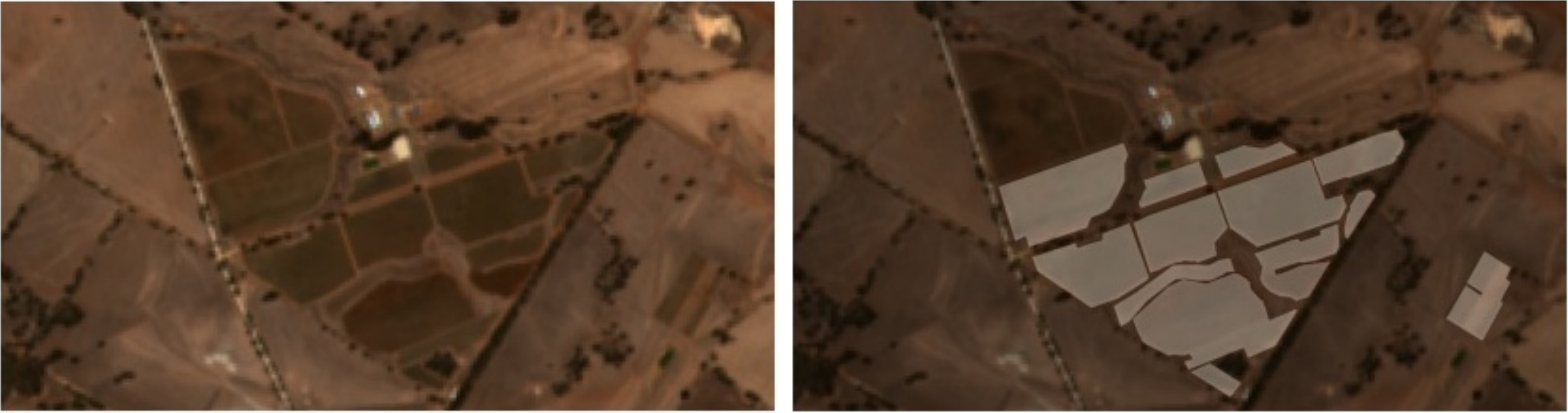}
    \caption{In this paper we compare methods for optimal utilisation of multi-spectral satellite images with CNNs. The left image shows the RGB channels of an image used to test the methods compared in this paper. The expert annotation of the target region (a vineyard) is also shown overlayed on the right.}
    \label{fig:val_mask_overlay}
\end{figure}

A number of works have successfully applied CNNs to satellite imagery, with various methods proposed to deal with the multi-band information \cite{Hamida2018}\cite{Li2017}\cite{Ribalta18}. However, due to a lack of consistent adoption of the same datasets, it is difficult to compare the performances of these methods. The purpose of this work is to provide a robust comparison of a number of representative techniques on the \textit{same data}. In this paper we use images of vineyards to train and evaluate our models, with the task being the semantic segmentation of the crop (grapes) from the background. An example image is given in Figure \ref{fig:val_mask_overlay}, with the RGB channels shown alongside an expert manual annotation.

In this work, we apply a standard deep learning architecture --- the U-Net \cite{Ronneberger2015} --- to the segmentation task, and specifically explore how the architecture can best utilise multi-spectral data. To explore how to best combine information in these bands, we compare the following methods: using bands as specified by a GIS expert; using all available bands; learning attention maps over the input bands; and leveraging Bayesian optimisation over the band choice. We demonstrate that simple methods can be used to significantly improve the performance of the model over expert band choice which, surprisingly, is still often preferred in industrial settings. We further show the performance boost afforded by the Bayesian optimisation, which to our knowledge is applied to band selection for the first time in this work.

The remainder of this paper is set out as follows: we first give a summary of the related work in this area in Section \ref{sec:related_work}, before describing the methods we compare for multi-spectral band selection applied to the task of satellite image segmentation in Section \ref{sec:methods}. Section \ref{sec:experimental_results} presents our experimental results, including a description of the data, implementation details, and assessment of the methods' performances.

\section{Related Work}
\label{sec:related_work}

A number of methods have been proposed for the task of band selection in the context of multi-spectral satellite image analysis, which can be broadly categorised into: manual methods; those based on classical probabilistic or information theory methods; and deep learning based techniques. 

One of the first works in this space, \cite{osti_674570}, describes various multi-spectral satellite bands, with the analysis providing heuristcs which can be used for manual band selection. In \cite{Band_Selection_pca}, the authors propose the use of a principal component analysis (PCA) based method for band selection, evaluating performances within a clustering framework. \cite{Feng2016} propose a probabilistic memetic search algorithm to identify the optimal bands for classification in satellite imagery.

Within a deep learning framework, \cite{audebert_2019} review various deep learning approaches for hyper-spectral classification. The processing of multi-band information has been tackled with CNNs both by alternating 1D (band-wise) and 2D (spatial) convolutions \cite{Hamida2018}, and with 3D convolutions \cite{Li2017}. \cite{Ribalta18} performed band selection from multi-spectral imagery using attention-based methods over all the available bands, focusing on the interpretability the method provides.

The CNN architecture we use as a backbone for our experiments is the widely adopted U-Net segmentation network \cite{Ronneberger2015}. The model has previously been applied to satellite imagery, for example \cite{AI4SG19} for Sentinel imagery and \cite{gaia} for DigitalGlobe data. We also note that we do not leverage the temporal dimension of satellite data, which could improve performance \cite{temporal}, but instead compare all methods with a simple but effective CNN architecture on static images.

To our knowledge, the closest related work to that detailed in this paper is \cite{Zhang19}, who compare a number of approaches for band selection for multi-spectral image classification with support vector machines (SVMs). Their compared methods include expert band selection and full selection of bands, with the study concluding that the inclusion of all bands outperforms expert selection in this case. 

In this paper, we compare a number of representative methods for band selection from multi-spectral imagery within a deep learning framework, evaluating all methods on the same data. The compared methods include: expert band selection; selection of all bands; an attention-based method; and Bayesian hyper-parameter optimisation for band selection.

\section{Methods}
\label{sec:methods}
In this section we describe the methods we compare on the task of multi-spectral satellite image segmentation. First, we discuss two forms of manual band selection: expert band selection --- which we use as a reference --- and providing all bands to the model. We then describe a model which uses an attention-based mechanism to select the optimal channel configuration, before concluding with a description of Bayesian hyper-parameter optimisation, which treats the selected channel indices as hyper-parameters.

\subsection{Manual Band Selection}

We first observe the protocol adopted in many industrial satellite image analysis tasks, and use expert domain knowledge to select the optimal bands for the task at hand. For this task, which involves the segmentation of vegetation, we use `Red', `Green' and `Infrared' channels. We also use `SWIR' (short-wave infrared) which is useful specifically for agricultural vegetation. The limitation of using domain knowledge is that it is often difficult to acquire and changes depending on the target application. Furthermore, utilisation of such a minimal set of bands may discard useful information and lower performance. 

We also run experiments with the opposite approach, in which we provide all bands as input to the model. The possible limitation with this method is that it encodes no inductive bias on the task and could lead to model overfitting. 

\subsection{Attention Over Channels}

We now present a method which explicitly learns attention over the channels of the input image, where we leverage the method used in `Squeeze-and-Excitation Networks' \cite{hu_2018}, in which attention over intermediate feature maps is used to improve performance. 

Intuitively, squeeze-excite blocks spatially pool all information in an input feature, producing an intermediate vector, before passing the result through a small neural network to produce a vector of values which act as weights (attention) over the input channels.

\begin{figure}
    \centering
    \includegraphics[width=\linewidth]{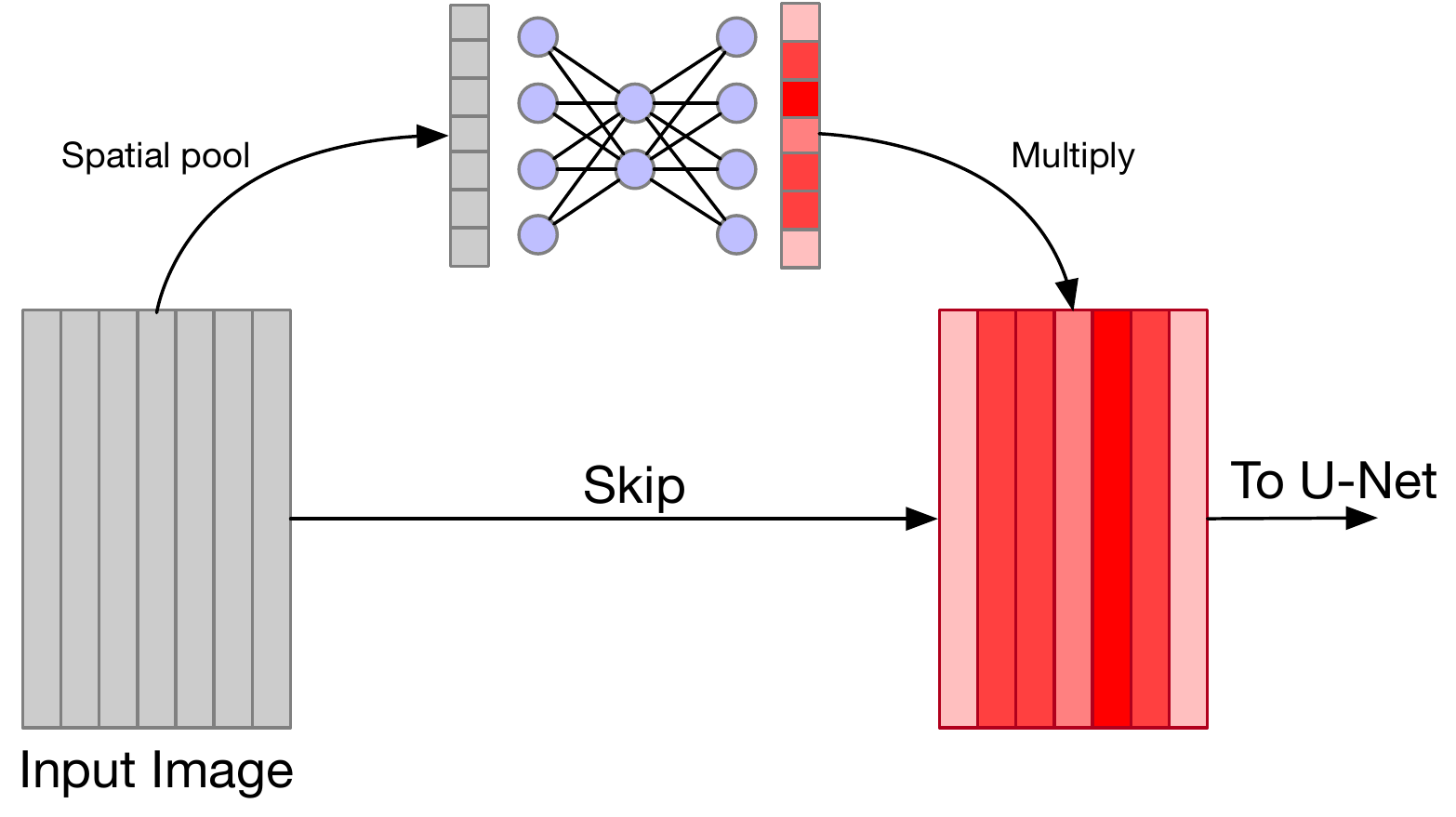}
    \caption{Illustration of a squeeze-excite block applied as an attention mechanism over a multi-spectral input image.}
    \label{fig:attention_block}
\end{figure}

Concretely, consider an input feature $\textit{\textbf{x}} \in \mathbb{R}^{H \times W \times C}$. The feature undergoes global average pooling to produce an intermediate vector, $\textit{\textbf{u}} \in \mathbb{R}^{1 \times 1 \times C}$, compressing all information in each channel to a single scalar value. This vector is then passed through a small multi-layer perceptron (MLP) with sigmoid activations on the output layer to produce the channel weights, $\textit{\textbf{h}} \in \mathbb{R}^{1 \times 1 \times C}$. The output representation, $\textit{\textbf{y}} \in \mathbb{R}^{H \times W \times C}$, is then computed by multiplying the input by the channel weights as: $\textit{\textbf{y}} = \textit{\textbf{x}} * \textit{\textbf{h}}$. The process is illustrated in Figure \ref{fig:attention_block}.

The use of an MLP to map global channel information to channel weights allows the mapping to be a non-linear combination of the input channels. We implement the MLP as a 2-layer neural network, where the intermediate representation introduces a bottleneck, such that it has $r=4$ times fewer neurons than the MLP's input, $\textit{\textbf{u}}$.  Differently to the Squeeze-and-Excitation networks, where these blocks are used to learn channel-wise dependencies over intermediate feature maps, we use the squeeze-excite block only on the multi-spectral input image. In this way, we learn attention over the input channels, effectively learning which channels contain the most salient information.

\subsection{Bayesian Hyper-parameter Optimisation}

The final band selection method with which we experiment is treating the choice of bands as a hyper-parameter to be tuned. Typically, hyper-parameters are tuned by training a number of models with varying hyper-parameter settings (varied by grid search) and choosing the optimal setting based on test or validation performance. For a small number of hyper-parameters, this is often sufficient. However, in this case, with a large number of bands, the grid search method becomes infeasible.

As an alternative, we experiment with the use of Bayesian hyper-parameter optimisation \cite{snoek_2012}. In this section, we briefly describe the Gaussian Process (GP), before outlining how it can be used for hyper-parameter optimisation and concluding by detailing how we use the method to choose the optimal band indices from multi-spectral data. 

\subsubsection{Gaussian Processes}

The idea behind Bayesian hyper-parameter optimisation is to place a Gaussian Process prior over the domain of the hyper-parameter in question. A GP is considered as a probability distribution over \textit{functions}, with the nature of these functions defined by a \textit{mean function}, $\mu(\textit{\textbf{x}})$, and a \textit{covariance kernel}, $k(\textit{\textbf{x}}, \textit{\textbf{x}}')$.  Here, the function over which we want to obtain a distribution is the mapping between the hyper-parameter setting, $\textit{\textbf{x}}$, and the model accuracy. In this work, we use the widely used Mat\'ern kernel as our covariance function and adopt common practise of a zero mean function.

\subsubsection{Optimisation through Gaussian Processes}

Gaussian Process hyper-parameter optimisation involves iteratively: (1) sampling a point from the domain of the hyper-parameter; (2) evaluating the model performance at this point; and (3) using the GP to estimate the best point from the domain to next sample, conditioning the estimate on the last and all previously sampled points. 

The process is conceptually simple, leaving only the question of the optimal way to leverage the GP to estimate the next point to sample. A number of methods have been proposed for this \cite{Wilson2018} --- in this work, we find the point which maximises the \textit{expected improvement} of the test loss under the GP model, as: 

\begin{equation}
    \textit{\textbf{x}}_{next} = \argmax_{\textit{\textbf{x}}} \mathbb{E} \left[ max(0, f_{best} - f(\textit{\textbf{x}})) \right]
\end{equation}

Here, $f_{best}$ represents the best test loss observed so far, and the expectation is optimised with Monte Carlo methods \cite{snoek_2012}.

\subsubsection{GPs for optimal band selection}

\begin{figure*}[t!]
    \centering
    \includegraphics[width=\textwidth]{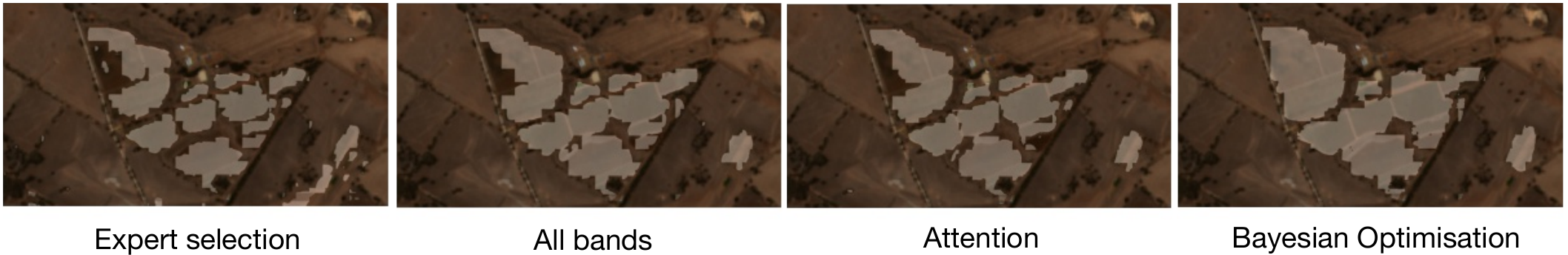}
    \caption{Predictions of the best model from each of the compared models on the test image.}
    \label{fig:preds}
\end{figure*}

In this paper, we use a GP to learn the mapping from the choice of channel indices of multi-spectral data to the CNN accuracy. Specifically, for each band, we introduce a binary variable $x_i \in \{0 , 1\}$, which represents whether or not the $i^{th}$ band is given as input to the CNN. Together, for $D$ possible bands to select from, the hyper-parameter to be tuned is $\textit{\textbf{x}} \in \{0 , 1\}^{D}$.

\section{Experimental Results}
\label{sec:experimental_results}

In this section we present our experimental results, including a description of the data used for the task and the experimental setup. Section \ref{sec:discussion} details both the qualitative and quantitative performances of the compared methods.

\subsection{Data}

For the experiments in this paper, we use data captured by the Sentinel-2 satellite, which captures multi-spectral data at a resolution of either 10m, 20m or 60m. The data is acquired via requests to the SentinelHub API \cite{sentinelhub}, which preprocesses the raw data before providing 12-band information given a specified geolocation. 

In this work, we tackle the task of vineyard (grape) segmentation. We take data from two different geolocations in South-West Australia (satellite images of two \textit{different} vineyards), using one location for training the model and another for testing. The training and test images both cover an area of 2.90 squared kilometres, and we request training images at 25 distinct timestamps from the year 2019 to increase the volume of the training data (requesting 2 timestamps for testing). We use the SentinelHub API to request data without cloud cover (maximum cover of 10\%) and normalise the data by the training images' mean and variance. To train the models described in this work, we obtain manual annotations of the target regions from a GIS expert. The test vineyard is shown in Figure \ref{fig:val_mask_overlay} alongside the expert annotation.

The images received from SentinelHub are too large to be processed at once due to GPU memory constraints. To overcome this, we break each image into a number of smaller tiles, passing each tile to the model in turn before reconstructing the tiles of the model's predictions. The resulting dataset consists of 11943 tiles for training, and 928 tiles for testing, each with dimensions $96 \times 96 \times 12$ (spatial dimensions by number of bands).

\subsection{Implementation Details}
\label{sec:implementation}

The backbone CNN architecture used for these experiments is the U-Net \cite{Ronneberger2015}), which provides near state-of-the-art results for a number of semantic segmentation tasks while retaining a simple \textit{Encoder - Decoder} structure.

We train the network on the 11943 training tiles with respect to the intersection-over-union score, which is a differentiable proxy for the Dice coefficient. The Dice coefficient is often of interest to end users of segmentation models as it is a reasonable measure of the qualitative similarity between the model prediction and ground truth. We adopt the Adam optimizer for gradient descent \cite{kingma2017} with an initial learning rate of $0.01$, using a mini-batch size of 128. For the manual band selection and attention based methods, we ensure our results are representative and reproducible by training models multiple (10) times with differing random seeds. For the Bayesian optimisation, we warm start the optimisation process by randomly sampling 5 points in the domain of $\textit{\textbf{x}}$, before taking 35 further samples and choosing the best 10 models. We note that each sample requires training and evaluating the CNN, a point upon which we elaborate in Section \ref{sec:discussion}. All models were trained on a single Tesla V100 for 25 epochs, which we observed as sufficient for the plateauing of training and evaluation performance. 

\subsection{Discussion}
\label{sec:discussion}

In this section we compare the performance of the following methods of band selection from satellite data: 
(1) manual selection of bands using expert knowledge;
(2) use of all available bands;
(3) learned attention over bands;
(4) selection of bands through Bayesian optimisation.

The Dice performances of our compared methods are given in Table \ref{dice_preds}. For the methods described in the top three rows, we show the statistics of performances of models trained from 10 random seeds. For band selection through Bayesian optimisation, we show the statistics of the 10 best performing hyper-parameter (band index) settings. 

Figure \ref{fig:preds} shows the predictions of the best performing model from each of the compared methods on the test vineyard. The qualitative results reasonably reflect the figures from Table \ref{dice_preds}. The expert band selection performs most poorly, with large areas of the target area omitted. Furthermore, use of all bands and the attention mechanism give visually similar results. Finally, we see that the best model returned by the Bayesian optimisation covers more of the target area than the other compared methods. 

\begin{table}[]
\centering
\begin{tabular}{l|c|}
\cline{2-2}
                                            & Dice Coefficient (\%) \\ \hline
\multicolumn{1}{|l|}{Expert band selection} & 81.15 $\pm$ 2.95          \\ \hline
\multicolumn{1}{|l|}{All bands selected}    & 84.72 $\pm$ 1.85          \\ \hline
\multicolumn{1}{|l|}{Attention over bands}  & 84.68 $\pm$ 1.81          \\ \hline
\multicolumn{1}{|l|}{Bayesian optimisation} & \textbf{86.53 $\pm$ 0.55} \\ \hline
\end{tabular}
\caption{Dice performance ($\mu \pm \sigma$) of the compared methods on the test images. Results are shown averaged over 10 models. For Rows 1-3, the 10 models are trained with 10 random seeds. For Row 4, the results represent the 10 best models from the Bayesian optimisation.}
\label{dice_preds}
\end{table}

We highlight that expert band selection has a significantly lower performance than the other compared methods. Though this is likely due to the additional information available to the other methods, we find the result salient as expert band selection is still standard practise in many industrial settings. Simply providing information from all bands to the model is sufficient to yield a 3.6\% improvement in accuracy, corroborating the work of \cite{Zhang19} with SVMs.

We also note that the introduction of attention over input channels has a negligible affect on the test time accuracy over the standard architecture. This can be explained as the convolution filters in the first layer of the of the standard architecture can learn to focus on specific channels of the input image. We conclude that the inductive bias encoded via the attention mechanism is not critical to the test performance.

Finally, we highlight the further performance boost provided by the Bayesian hyper-parameter optimisation. The method provides an average 1.9\% Dice improvement over methods (2) and (3), and over 5\% improvement over expert band selection. This method does, however, come with significant computational cost. The Gaussian Process optimisation was run for 40 iterations, each of which required training and evaluating the CNN. For the models discussed in this paper, a standard U-Net architecture trained on a dataset containing 12K images, each iteration took roughly 18 minutes. As a result, a single run of GP optimisation took 13 hours on a high performance GPU. Given that GP parameters may themselves need tuning, and that many datasets are far larger than the one demonstrated upon in this work, the accuracy boost afforded by this method may be deemed insufficient for the extra computation required, depending on the task at hand.

\section{Conclusion}

In this work, we have compared methods for optimal utilisation of multi-spectral information in satellite imagery, applied to land cover segmentation with the example of vineyard detection. We compared four representative methods of band selection in the context of deep learning models, evaluating their performances on the same data. We have explored two forms of manual band selection: expert band selection, which we use as a reference, and providing all bands to the model. We also compare a model which uses an attention-based mechanism to learn the most salient bands, and Bayesian hyper-parameter optimisation which we apply to band selection for the first time in this work. Our experiments demonstrated that using all available band information significantly outperforms expert band selection and show that the Bayesian optimisation can be leveraged to further boost performance to 5\% over expert selection.

{
\bibliographystyle{named}
\bibliography{AI4GoodHarvard.bib}

\begin{thebibliography}{}

\bibitem[\protect\citeauthoryear{Andries \bgroup \em et al.\egroup
  }{2019}]{andries2019seeing}
Ana Andries, Stephen Morse, Richard~J Murphy, Jim Lynch, and Emma~R Woolliams.
\newblock Seeing sustainability from space: Using earth observation data to
  populate the un sustainable development goal indicators.
\newblock {\em Sustainability}, 11(18):5062, 2019.

\bibitem[\protect\citeauthoryear{Audebert \bgroup \em et al.\egroup
  }{2019}]{audebert_2019}
Nicolas Audebert, Bertrand Saux, and S{\'{e}}bastien Lef{\`{e}}vre.
\newblock {Deep Learning for Classification of Hyperspectral Data: A
  Comparative Review}.
\newblock {\em IEEE Geoscience and Remote Sensing Magazine}, 7(2):159--173, apr
  2019.

\bibitem[\protect\citeauthoryear{{Ben Hamida} \bgroup \em et al.\egroup
  }{2018}]{Hamida2018}
A.~{Ben Hamida}, A.~{Benoit}, P.~{Lambert}, and C.~{Ben Amar}.
\newblock 3-d deep learning approach for remote sensing image classification.
\newblock {\em IEEE Transactions on Geoscience and Remote Sensing},
  56(8):4420--4434, 2018.

\bibitem[\protect\citeauthoryear{Clodius \bgroup \em et al.\egroup
  }{1998}]{osti_674570}
W.B. Clodius, P.G. Weber, C.C. Borel, and B.W. Smith.
\newblock Multi-spectral band selection for satellite-based systems.
\newblock 9 1998.

\bibitem[\protect\citeauthoryear{{DigitalGlobe}}{2019}]{dig_globe_report}
{DigitalGlobe}.
\newblock Seeing sustainability from space: Using earth observation data to
  populate the un sustainable development goal indicators.
\newblock
  \url{https://dgv4-cms-production.s3.amazonaws.com/uploads/document/file/169/DG_UNSDG_eBook_1-7-2019_digital_final.pdf},
  2019.

\bibitem[\protect\citeauthoryear{Efremova \bgroup \em et al.\egroup
  }{2019}]{AI4SG19}
Natalia Efremova, Dennis West, and Dmitry Zausaev.
\newblock Ai-based evaluation of the sdgs: The case of crop detection with
  earth observation data.
\newblock {\em CoRR}, abs/1907.02813, 2019.

\bibitem[\protect\citeauthoryear{Feng \bgroup \em et al.\egroup
  }{2016}]{Feng2016}
Liang Feng, Ah-Hwee Tan, Meng-Hiot Lim, and Si~Wei Jiang.
\newblock {Band selection for hyperspectral images using probabilistic memetic
  algorithm}.
\newblock {\em Soft Computing}, 20(12):4685--4693, 2016.

\bibitem[\protect\citeauthoryear{Hu \bgroup \em et al.\egroup }{2018}]{hu_2018}
Jie Hu, Li~Shen, and Gang Sun.
\newblock {Squeeze-and-Excitation Networks}.
\newblock In {\em Proceedings of the IEEE Computer Society Conference on
  Computer Vision and Pattern Recognition}, pages 7132--7141. IEEE Computer
  Society, dec 2018.

\bibitem[\protect\citeauthoryear{Jones \bgroup \em et al.\egroup }{2020}]{gaia}
Eriita~G. Jones, Sebastien Wong, Anthony Milton, Joseph Sclauzero, Holly
  Whittenbury, and Mark~D. McDonnell.
\newblock The impact of pan-sharpening and spectral resolution on vineyard
  segmentation through machine learning.
\newblock {\em Remote Sensing}, 12(6), 2020.

\bibitem[\protect\citeauthoryear{Kingma and Ba}{2015}]{kingma2017}
Diederik~P. Kingma and Jimmy~Lei Ba.
\newblock {Adam: A method for stochastic optimization}.
\newblock In {\em 3rd International Conference on Learning Representations,
  ICLR 2015 - Conference Track Proceedings}. International Conference on
  Learning Representations, ICLR, dec 2015.

\bibitem[\protect\citeauthoryear{Koonsanit \bgroup \em et al.\egroup
  }{2012}]{Band_Selection_pca}
Kitti Koonsanit, Chuleerat Jaruskulchai, and Apisit Eiumnoh.
\newblock Band selection for hyperspectral imagery with pca-mig.
\newblock In Zhifeng Bao, Yunjun Gao, Yu~Gu, Longjiang Guo, Yingshu Li, Jiaheng
  Lu, Zujie Ren, Chaokun Wang, and Xiao Zhang, editors, {\em Web-Age
  Information Management}, pages 119--127, Berlin, Heidelberg, 2012. Springer
  Berlin Heidelberg.

\bibitem[\protect\citeauthoryear{{Li} \bgroup \em et al.\egroup
  }{2017}]{Li2017}
Y.~{Li}, H.~{Zhang}, and Q.~{Shen}.
\newblock Spectral–spatial classification of hyperspectral imagery with 3d
  convolutional neural network.
\newblock {\em Remote Sensing}, 9(1):67, 2017.

\bibitem[\protect\citeauthoryear{Lorenzo \bgroup \em et al.\egroup
  }{2018}]{Ribalta18}
Pablo~Ribalta Lorenzo, Lukasz Tulczyjew, Michal Marcinkiewicz, and Jakub
  Nalepa.
\newblock Band selection from hyperspectral images using attention-based
  convolutional neural networks.
\newblock {\em CoRR}, abs/1811.02667, 2018.

\bibitem[\protect\citeauthoryear{Research and
  Markets}{2020}]{research_and_markets2020}
Research and Markets.
\newblock Geographic information system market worth \$14.5 billion by 2025.
\newblock {\em technical report}, 2020.

\bibitem[\protect\citeauthoryear{Ronneberger \bgroup \em et al.\egroup
  }{2015}]{Ronneberger2015}
Olaf Ronneberger, Philipp Fischer, and Thomas Brox.
\newblock {U-Net: Convolutional Networks for Biomedical Image Segmentation}.
\newblock {\em MICCAI}, 2015.

\bibitem[\protect\citeauthoryear{SentinelHub}{}]{sentinelhub}
SentinelHub.
\newblock {SentinelHub API}.

\bibitem[\protect\citeauthoryear{Snoek \bgroup \em et al.\egroup
  }{2012}]{snoek_2012}
Jasper Snoek, Hugo Larochelle, and Ryan~P. Adams.
\newblock {Practical Bayesian optimization of machine learning algorithms}.
\newblock In {\em Advances in Neural Information Processing Systems}, volume~4,
  pages 2951--2959, jun 2012.

\bibitem[\protect\citeauthoryear{{United
  Nations}}{2015}]{united2015transforming}
{United Nations}.
\newblock Transforming our world: The 2030 agenda for sustainable development.
\newblock {\em General Assembley 70 session}, 2015.

\bibitem[\protect\citeauthoryear{Vuolo \bgroup \em et al.\egroup
  }{2018}]{temporal}
Francesco Vuolo, Martin Neuwirth, Markus Immitzer, Clement Atzberger, and
  Wai-Tim Ng.
\newblock How much does multi-temporal sentinel-2 data improve crop type
  classification?
\newblock {\em International Journal of Applied Earth Observation and
  Geoinformation}, 72:122--130, 10 2018.

\bibitem[\protect\citeauthoryear{Wilson \bgroup \em et al.\egroup
  }{2018}]{Wilson2018}
James~T. Wilson, Frank Hutter, and Marc~Peter Deisenroth.
\newblock {Maximizing acquisition functions for Bayesian optimization}.
\newblock {\em Advances in Neural Information Processing Systems},
  2018-Decem:9884--9895, may 2018.

\bibitem[\protect\citeauthoryear{Zhang \bgroup \em et al.\egroup
  }{2019}]{Zhang19}
Tian-Xiang Zhang, Jin-Ya Su, Cun-Jia Liu, and Wen-Hua Chen.
\newblock Potential bands of sentinel-2a satellite for classification problems
  in precision agriculture.
\newblock {\em International Journal of Automation and Computing}, 16/1, 2019.

\end{thebibliography}
}

\end{document}